\documentclass[letterpaper, 10 pt, conference]{ieeeconf}  

\IEEEoverridecommandlockouts                              

\usepackage{graphicx} 
\usepackage{amsmath} 
\usepackage{amssymb}  
\usepackage{float}
\usepackage[acronym]{glossaries}
\usepackage{color}
\usepackage{cite}
\usepackage{subcaption}
\usepackage[font=small,labelfont=bf,tableposition=top]{caption}
\definecolor{dgreen}{rgb}{0,0,0}
\definecolor{dyellow}{rgb}{.7,.7,0}
\definecolor{dred}{rgb}{1,0,0}
\definecolor{dblue}{rgb}{0,0,0.7}
\definecolor{dorange}{rgb}{0.9,0.5,0.1}

\newacronym{vapo}{VAPO}{Visual Affordance-guided Policy Optimization}
\makeatletter
\let\NAT@parse\undefined
\makeatother

\usepackage[hidelinks]{hyperref}
\hypersetup{
    colorlinks,
    urlcolor=magenta,
    citecolor={black},
    linkcolor={black}
    }

\title{\LARGE \bf
Affordance Learning from Play for Sample-Efficient Policy Learning
}

\author{Jessica Borja-Diaz$^\ast$, Oier Mees$^\ast$, Gabriel Kalweit, Lukas Hermann, Joschka Boedecker, Wolfram Burgard\\
\url{http://vapo.cs.uni-freiburg.de}
\thanks{$^\ast$Equal contribution. All authors are with the University of Freiburg, Germany. This work has  been supported partly by the German Federal Ministry of Education and Research under contract 01IS18040B-OML}
}

\begin{document}

\makeatletter
\let\@oldmaketitle\@maketitle
\renewcommand{\@maketitle}{\@oldmaketitle
  \includegraphics[width=0.5\linewidth]
    {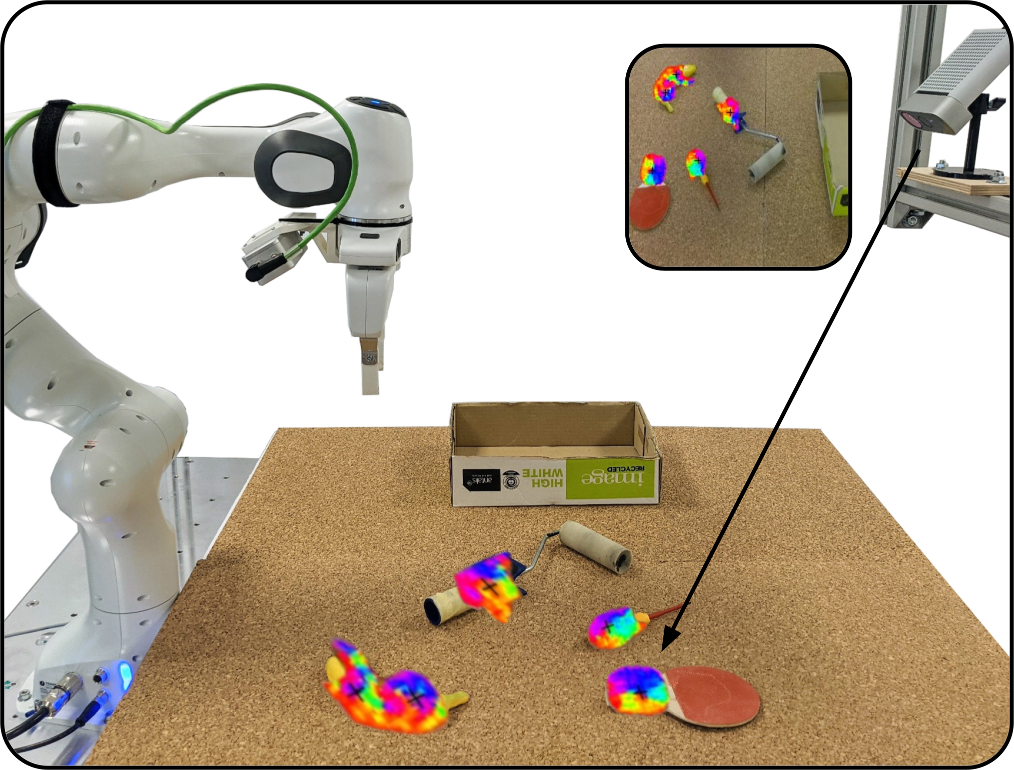}
  \includegraphics[width=0.5\linewidth]
    {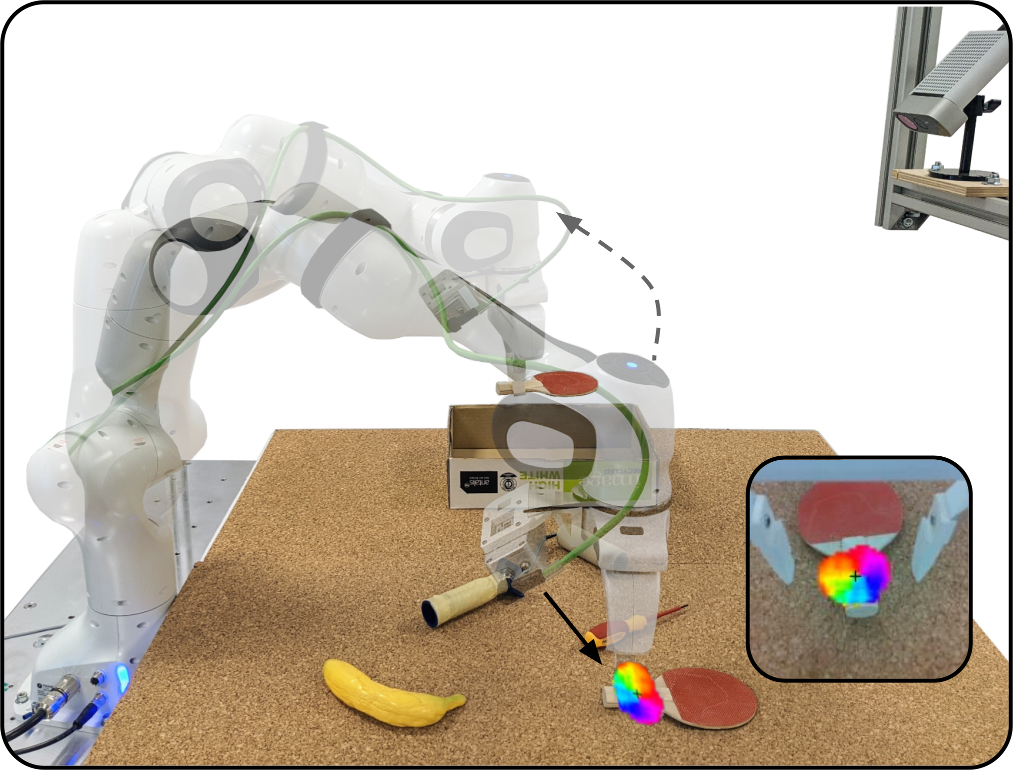} \\[0.25em]
  \refstepcounter{figure}\footnotesize{\bf{Fig. 1:} Real world setup for a tidy up task: our self-supervised visual affordance model guides the robot to the vicinity of actionable regions in the environment with a model-based policy.  Once inside this area, we switch to a local reinforcement learning policy, in which we embed our affordance model to favor the same object regions favored by people and to boost sample-efficiency.}
  \label{fig:real} \medskip \vspace{-10pt}}
\makeatother

\maketitle
\thispagestyle{empty}
\pagestyle{empty}

\renewcommand\thefigure{\arabic{figure}}
\setcounter{figure}{1}
\begin{abstract}
    Robots operating in human-centered environments should have the ability to understand how objects function: \emph{what} can be done with each object, \emph{where} this interaction may occur, and \emph{how} the object is used to achieve a goal. 
    To this end, we propose a novel approach that extracts a self-supervised visual affordance model from human teleoperated play data and leverages it to enable efficient policy learning and motion planning.
    We combine model-based planning with model-free deep reinforcement learning (RL) to learn policies that favor the same object regions favored by people, while requiring minimal robot interactions with the environment.
    We evaluate our algorithm, Visual Affordance-guided Policy Optimization (VAPO), with both diverse simulation manipulation tasks and real world robot tidy-up experiments to demonstrate the effectiveness of our affordance-guided policies.
    We find that our policies train $4 \times$ faster than the baselines and generalize better to novel objects because our visual affordance model can anticipate their affordance regions.

\end{abstract}



\section{Introduction}
    Humans have the ability to effortlessly recognize and infer functionalities of objects despite their large variation in appearance and shape. For example, we understand that we need to pull the handle of a drawer to open it or grasp a knife by the handle to use it.
    This capacity to focus on the most relevant behaviors in a given situation enables efficient decision making by limiting the choices of action that are even considered. Gibson's theory of affordances~\cite{gibson1979ecological} provides a way to reason about object function. It suggests that objects have action possibilities, e.g., a mug is ``graspable'' and a door is ``openable'' and has been extensively studied in both the robotics and the computer vision communities~\cite{hassanin2018visual}.
    
    However, the abstract notion of ``what actions are possible?'' addressed by current affordance learning methods is limited. A robot needs to know \emph{where} are actionable regions in an environment,  the specific points on the object that need to be manipulated for a successful interaction, \emph{what} it can achieve with it and \emph{how} the object is used to achieve a goal. Current affordance learning methods have two major problems. First, they are limited by requiring heavy supervision in the form of manually annotated segmentation masks~\cite{nguyen2016detecting, myers2015affordance, nguyen2017object, do2018affordancenet} or expensive interactive exploration~\cite{mo2021where2act, nagarajan2020learning}, restricting their scalability and applicability in practical robotics scenarios. Second, current affordance-augmented robotic systems are limited in the complexity of the actions they model by relying often on predefined action templates~\cite{mo2021where2act, nagarajan2020learning, zeng2018robotic, yen2020learning}. Together, these limitations naturally restrict the scope of affordance learning systems to a narrow set of objects and robotics applications.
    
    In light of these issues, we propose a method for sample-efficient policy learning of complex manipulation tasks that is guided by a self-supervised visual affordance model. Therefore, we call our algorithm \gls{vapo}. Towards overcoming the issues of expensive manual supervision and  exploration, we propose to learn affordances that are \emph{grounded} in real human behavior from teleoperated play data~\cite{lynch2020learning}. Play data is not random, but rather structured by human knowledge of object affordances (e.g., if people see a drawer in a scene, they tend to open it). Moreover, affordances discovered from unlabeled play are \emph{functional affordances}, priming a robot to approach an object the way a human would. On the other hand, teleoperated play data does not bear the risk of the correspondence problem as opposed to recordings directly from human demonstrations. We hence leverage this visual affordance model to guide a robot to perform complex manipulation tasks. Aside from accelerating learning, a critical advantage of imbuing robots with an object-centric visual affordance prior is generalization: the learned policy generalizes to unseen object instances because our visual affordance model can anticipate their affordance regions.
    
    Our approach decomposes object manipulation into a sample-efficient combination of model-based planning and model-free reinforcement learning, inspired by a recent line of work that aims to combine classical motion planning with machine learning~\cite{silver2018residual, lee2020guided,ichter2020broadly}. Concretely, we first predict object affordances and drive the end-effector from free-space to the vicinity of the afforded region with a model-based method. Once inside this area, the model cannot be trusted and we switch to a RL policy in which the agent is rewarded for interacting with the afforded regions. This way, the local policy has a ``human prior'' for how to approach an object, but is free to discover its exact grasping strategy. Our self-supervised visual affordance model is leveraged twice to boost sample-efficiency: 1) driving the model-based planner to the vicinity of afforded regions, 2) guiding a local grasping RL policy to favor the same object regions favored by people. Standard model-free RL faces a number of challenges, since the policy must solve two problems: representation learning and task learning from high-dimensional raw observations in a single end-to-end training procedure. As in practice solving \emph{both} problems together is difficult, embedding our visual affordance model within a reinforcement learning loop alleviates the representation learning challenge. The interplay between model-based and model-free policies allows for a sample-efficient division of the robot control learning, without assuming a predefined set of manipulation primitives, 3D object shapes or a tracking system.

\section{Related Work}

\textbf{Predicting Semantic Representations} 
To successfully interact with a 3D object, a robot must be able to ``understand'' it given some perceptual observation. There exists a large body of work in the computer vision community targeting such an understanding in the form of different semantic labels. For example, predicting category labels~\cite{chang2015shapenet}, or more fine-grained output such as semantic keypoints~\cite{you2020keypointnet}, part segmentations~\cite{mo2019partnet} or afforded spatial relations~\cite{mees20icra_placements} can arguably yield more actionable representations e.g. allowing one to infer where ``handles'' are. However, merely obtaining such semantic labels is clearly not sufficient on its own, a robot must also understand \emph{what} needs to be done and \emph{how}  the object is used to achieve a goal.

\textbf{Acting with Model-based Planning}
Towards obtaining useful information for how to act, some methods aim for representations that can be leveraged by classical robotics techniques. In particular, traditional analytical approaches use knowledge of the 3D object pose~\cite{xiang2017posecnn, tremblay2018deep}, shape~\cite{chang2015shapenet, mees19iros}, gripper configuration, friction coefficients, etc. to determine  optimal action trajectories. However, model-based methods rely on an accurate model of the environment and they normally do not handle perception errors and physical interactions naturally~\cite{mees21iser}, limiting their reliability. Our approach uses model-based planning to guide the robot to the vicinity of detected affordance regions and switches then to a local RL policy.

\textbf{Reinforcement Learning Grasping}
RL models offer a counterpoint to the planning paradigm. Instead of breaking the task into two steps, static grasp synthesis followed by motion planning, it can operate directly from raw sensory inputs in closed-loop feedback control, which are not subject to estimation errors~\cite{levine2016end,kalashnikov2018qt}. Unlike model-based methods, RL methods do not require a detailed description of the environment and the task, but rather require access to interaction with the environment and to a reward function. Such binary rewards are easy to describe, but unfortunately they render RL methods extremely sample-inefficient and brittle. Although there have been promising advances in learning data-driven reward functions~\cite{Sermanet2017TCN, singh2019, mees20icra_asn}, for most complex problems of interest, learning RL policies from scratch remains intractable. In contrast, we inject an object-centric visual affordance prior extracted from human teleoperated play data to boost sample efficiency.

\textbf{Visual Affordances}
Most closely related to our approach is the line of work where visual affordances are learned for object manipulation~\cite{lenz2015deep, levine2018learning, do2018affordancenet, nguyen2017object, mandikal2020dexterous}. Traditionally, visual affordance learning methods are limited by their requirement of manually drawn segmentation masks or keypoints~\cite{nguyen2016detecting, myers2015affordance, nguyen2017object, do2018affordancenet} and some leverage additional sensing, such as force gloves~\cite{castellini2011using}. Recently, there has been a shift to explore other forms of supervision such as videos~\cite{nagarajan2019grounded},  a robot's gripper grasp success/failure~\cite{yen2020learning,lenz2015deep} or thermal image contact data~\cite{mandikal2020dexterous}. In contrast, we leverage a self-supervised signal of a robot's gripper opening and closing during human teleoperation to learn image-based functional affordances.

\begin{figure*}[t]
    \centering
    \begin{tabular}{cc}
        \includegraphics[width=0.7\linewidth]{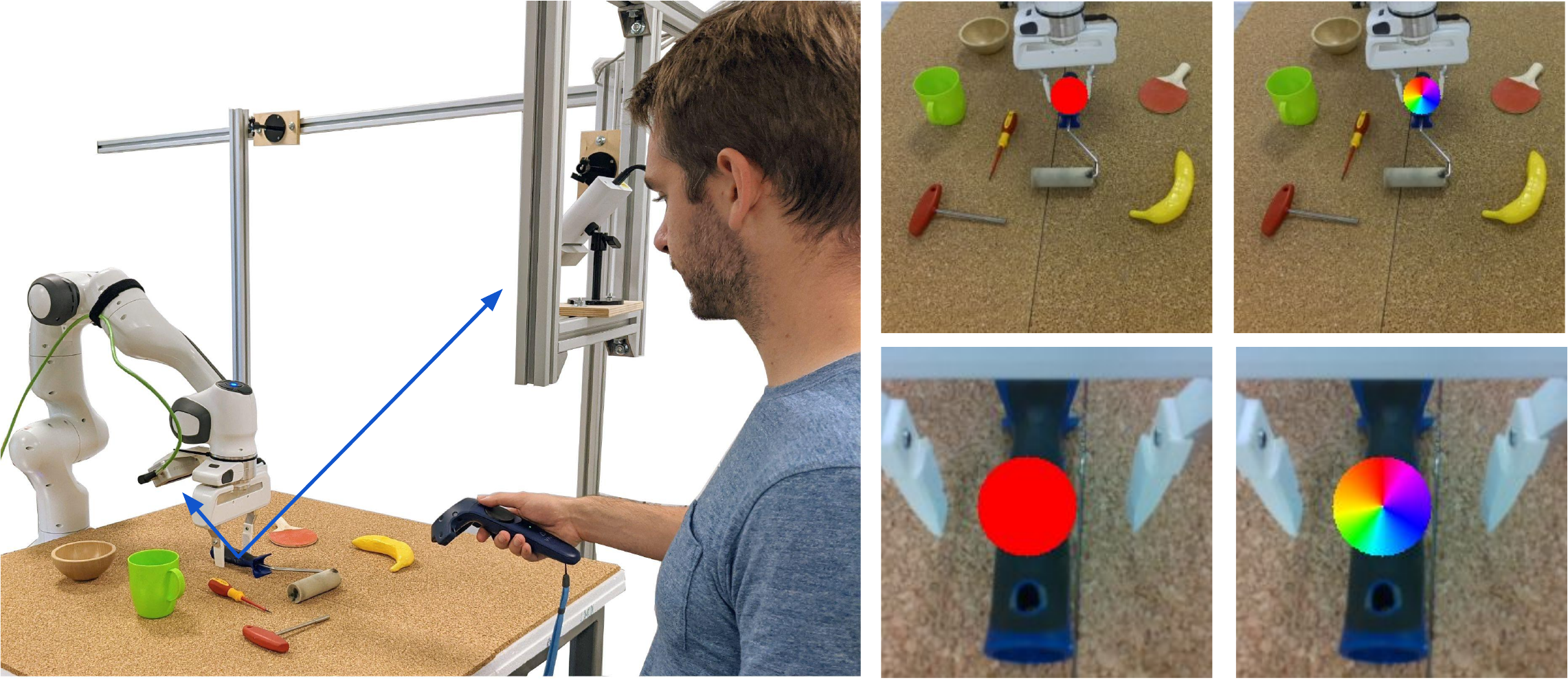} &
        \includegraphics[width=0.15\linewidth]{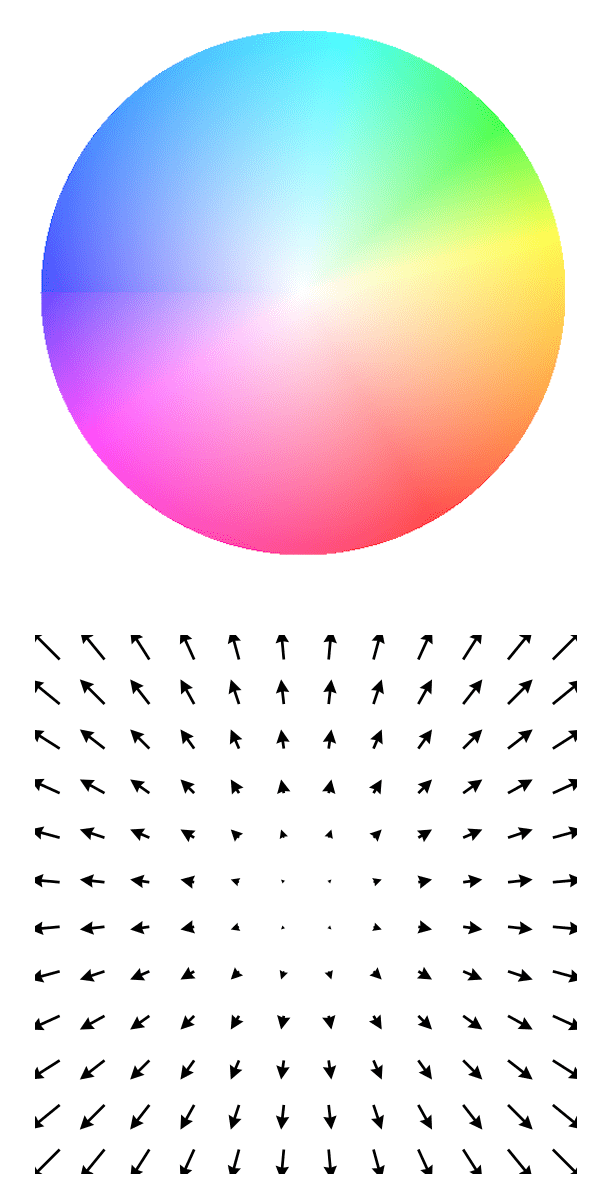}
    \end{tabular}
     \caption{Visualization of our self-supervised object affordance labelling.  We leverage a self-supervised signal of a robot's gripper opening and closing during human teleoperation to project the 3D tool-center-point into the static and gripper cameras. We label the neighboring pixels within a radius around the afforded region with a binary segmentation mask and direction vectors from each pixel towards the affordance region center. On the right we show the color code used to interpret the direction vectors.}
    \label{fig:labeling}
\end{figure*}

\section{Approach}
The main incentive of our method is to learn sample-efficient policies of complex manipulation tasks that are guided by a self-supervised visual affordance model. Our approach consists of three steps. First, we train a network to discover and learn object affordances in unlabeled play data (Sec. \ref{sec:affordance_play}). Second, we divide the space into regions where a model-based policy is reliable and regions where it may have limitations handling perception errors or physical interactions. We leverage the learned affordance model to drive the end-effector from free-space to the vicinity of the afforded region with a model-based policy $\pi_{mod}$ (Sec. \ref{sec:mb_to_rl}). Third, once inside this area we switch to a local reinforcement learning policy $\pi_{rl}$, in which we embed our affordance model to favor the same object regions favored by people and to boost sample-efficiency (Sec. \ref{sec:rl}). Thus, our final policy is defined as a mixture:
\begin{equation}
    \pi(a|s) = (1 - \alpha(s)) \cdot \pi_{mod} (a|s) + \alpha(s) \cdot \pi_{rl}(a|s),
    \label{eq:policy}
\end{equation}
where $\alpha(s)\in[0,1]$. We use an estimate of the normalized distance between the robot's gripper and the affordance region $\alpha(s)$ to switch between the policies. An overview of the system is given in Figure~\ref{fig:real}.

\subsection{Learning Visual Affordances from Play}
\label{sec:affordance_play}

Our key insight is to learn about object interactions from play data by leveraging a self-supervised signal of a robot's gripper opening and closing during human teleoperation, as shown in Figure~\ref{fig:labeling}. In this way, without explicit manual segmentation labels, we learn to anticipate not only \emph{where} are regions that afford human-object interactions, but also a powerful prior on \emph{how} humans approach those objects. The only assumption our method makes is an existing robot-camera calibration. We decouple the affordance prediction task into different components.

First, the affordance model $\mathcal{F}_a$ learns to transform an image $I$  into a binary segmentation map $A \in \mathbb{R}^{H \times W}$, indicating regions that afford an interaction. Second, it estimates 2D pixel coordinates of the affordance region centers by predicting a vector from each affordance pixel towards the center. Estimating the center points of the afforded regions is a key component in order to disambiguate affordances from multiple objects in a scene.
Clearly, play data showing people naturally interacting with objects partially reveals the afforded regions in an environment. Thus, in order to discover affordances in unlabeled data the gripper action is used as a heuristic to detect human-object interactions.

Intuitively, if the gripper closes during play, it is indicative of a possible interaction that will start at that position. Thus, we can project the gripper's 3D point $p^t_{grip}$ to a camera image pixel $u^t_{grip}$ and label the pixels within a radius $r$ for the past $n$ frames as an afforded region. Similarly, if the gripper transitions from being closed to open, it means that an interaction with an object ended at the 3D position $p_{i}$. This allows us to discover a set  of interaction points throughout time $P^k = (p_1, p_2, ..., p_k)$, which represent the world coordinates of where interactions have occurred until timestep $k$.
To get the full set of interaction locations for a timestep $t$ we consider the 3D positions from where a grasp will occur and where an interaction has been previously occurred until $t$. Finally, each 3D point is projected to a camera image pixel to create the affordance mask label by marking neighboring pixels. The pixel coordinates of the projected points are used as the affordance region centers.

In order to disambiguate affordances from multiple objects in a scene, we let the network estimate 2D pixel coordinates of the affordance region centers by predicting a vector from each affordance pixel towards the center $V  \in \mathbb{R}^{H \times W \times 2}$. We construct these labels by calculating the displacement from each pixel of the affordance mask to the corresponding projected center. The background pixels are pointed towards a fixed position to avoid false positives.

One limitation of the proposed heuristic is that it assumes users interacting with the environment during play will only close the gripper to perform meaningful interactions. To avoid erroneous labeling due to closing/opening the gripper in free-space without object-interaction, we introduce an additional check by requiring the griper-width to stay for $\Delta t$ timesteps in a range between opened and closed.

To train the full affordance model $F_a$ we apply two different loss functions. For the affordance segementation $A$ loss , we use a weighted sum between a cross entropy $\ell_{ce}$ and a dice loss $\ell_{dice}$ to account for class imbalance. 
Similar to Xie \emph{et al.}~\cite{xie2021unseen}, for the direction prediction we optimize a weighted cosine similarity loss given by:

\begin{equation*}
    \ell_{dir} = \sum_{i \in \mathcal{O}}{ 
        \alpha_i(1 - V^{\textbf{T}}_i \bar{V}_i) +
        \frac{\lambda_{b}}{\mid \mathcal{B} \mid}
        \sum_{i \in \mathcal{B}}{
            \left(
            1 - V^{\textbf{T}}_i
            \begin{bmatrix}
            0 \\
            1
            \end{bmatrix}
            \right)
        }
    }
\end{equation*}

Where $V_i$, $\bar{V}_i$ are the predicted and ground truth unit directions of pixel $i$ respectively. $\mathcal{B}, \mathcal{O}$ are the sets of pixels belonging to the background and affordance region classes. The total loss for the affordance model is given by $w_{ce} \ell_{ce} + w_{dice} \ell_{dice} + w_{dir} \ell_{dir}$.

\subsection{From Model-Based to Reinforcement Learning Workspace}
\label{sec:mb_to_rl}
Classical motion planning algorithms have difficulty in the presence of stochastic dynamics and high-dimensional systems. RL methods on the other hand offers a promising solution for its ability to learn general policies that can handle complex interactions and high-dimensional observations. However, for most complex problems of interest, learning RL policies from scratch remains intractable. Inspired by recent works that aim to combine both type of controllers~\cite{silver2018residual, lee2020guided,ichter2020broadly}, we divide the space into regions where a model-based policy is reliable and regions where it may have limitations handling perception errors or physical interactions. 

Concretely, we predict affordances and the corresponding region centers using a static camera image. Given this information of \emph{where} are regions that afford human-object interactions, we localize a chosen pixel region center in 3D and drive the end-effector from free-space to the vicinity of the afforded region with a model-based policy $\pi_{mod}$ and hand control over to the model-free policy~$\pi_{rl}$. We use an estimate of the distance between the robot's gripper and the predicted affordance region center to switch between the policies. Restricting the area where the RL policy is active to the vicinity of regions that afford human-object interactions has the advantage that it makes it more sample-efficient. Besides, this division of labour allows to learn local RL policies by switching to a gripper camera, improving generalization across different locations.

\subsection{Affordance-guided Reinforcement Learning Grasping}
\label{sec:rl}
Once the model-based policy $\pi_{mod}$ has brought the end-effector to the vicinity of a region that affords human-object interactions, we switch to a local gripper-camera based RL policy which we augment with an object-centric visual affordance prior to boost sample efficiency.

\textbf{Problem Formulation}: We consider the standard Markov decision process (MDP) $M = (\mathcal{S}, \mathcal{A}, \mathcal{T}, r, \mu_0, \gamma)$, where $\mathcal{S}$ and $\mathcal{A}$ denote the state space and action space respectively. 
$\mathcal{T} (s'| s, a)$ is the probability of transitioning from state $s$ to state $s'$ when applying action $a$. The actions are drawn from
a probability distribution over actions $\pi(a|s)$ referred to as the agent's policy.
$r(s, a)$ is the reward received by an agent for executing action $a$ in state $s$, $\mu_0$ the initial state distribution, and $\gamma \in (0, 1)$ the discount factor prioritizing long- versus short-term reward. The goal in RL is to optimize a policy $\pi(a | s)$ that maximizes the expected discounted return $\mathbb{E}_{\pi, \mu_0, \mathcal{T}}  \left [ \sum_{t=0}^{\infty} \gamma^t r(s, a) \right ]$.

\textbf{Observation Space}:
The observation space is composed of two parts: 1) the proprioceptive state including the 3D world coordinates of the end effector, the orientation euler angles and the gripper width. 2) The visual inputs consisting of the current RGB-D image observed by the gripper camera and the binary affordance mask predicted from the corresponding affordance model.

\textbf{Action Space}:  We use a 7-DOF Franka Emika Panda robot with a parallel gripper both in simulation and in the real world. The action space consists of delta XYZ position, delta euler angles and the binary gripper action.

\textbf{Reward}:
The reward function should not only signal a successful object interaction, but also guide the exploration process to focus on actionable object regions. To realize this,  we leverage the visual affordance model to guide the agent to get close to the affordance centers. This way, the local policy has a ``human prior'' for how  to  approach  an  object,  but  is  free  to  discover  its  exact grasping strategy.  Given the detected affordance center and the fact that the RL policy only acts locally within a neighborhood, we normalize the euclidean distance between the end effector and the affordance center to create a positive reward $R_{aff}$ which increases as we get closer to the detected center. Additionally if the agent goes outside the neighborhood, it receives a negative reward $R_{out}$ and if it successfully manipulates an object it receives a positive reward of $R_{succ}$. Our total reward function is:

\begin{equation}
    r(s,a) = \lambda_1 R_{succ} + \lambda_2 R_{aff} + \lambda_3 R_{out}
\end{equation}

\section{Implementation Details}
\textbf{Teleoperated play data}
During the unscripted  teleoperated interactions we record images from two cameras: a static camera that captures the global scene, and a camera mounted on the robot's gripper. The static camera image has a resolution of $200 \times 200$ and the gripper camera uses a resolution of $64 \times 64$. We label the images of the static camera with a radius of $r=10$ pixels around the projected center and the the gripper camera images with $r=25$.

\textbf{Affordance model}
 We use a U-net \cite{ronneberger2015unet} architecture followed by two parallel branches of convolutional layers that produce the affordance mask and center directions. Similar to Xiang \emph{et al.}~\cite{xiang2018posecnn}, we use a Hough voting layer to predict the 2D object centers during inference. The Hough voting layer takes the affordance mask and the direction vectors as input to compute a score for each pixel, indicating its likelihood of being an affordance region center. The location with the maximum score is selected as the object center. 

\begin{figure}[t]
    \centering
    \includegraphics[width=0.47\textwidth]{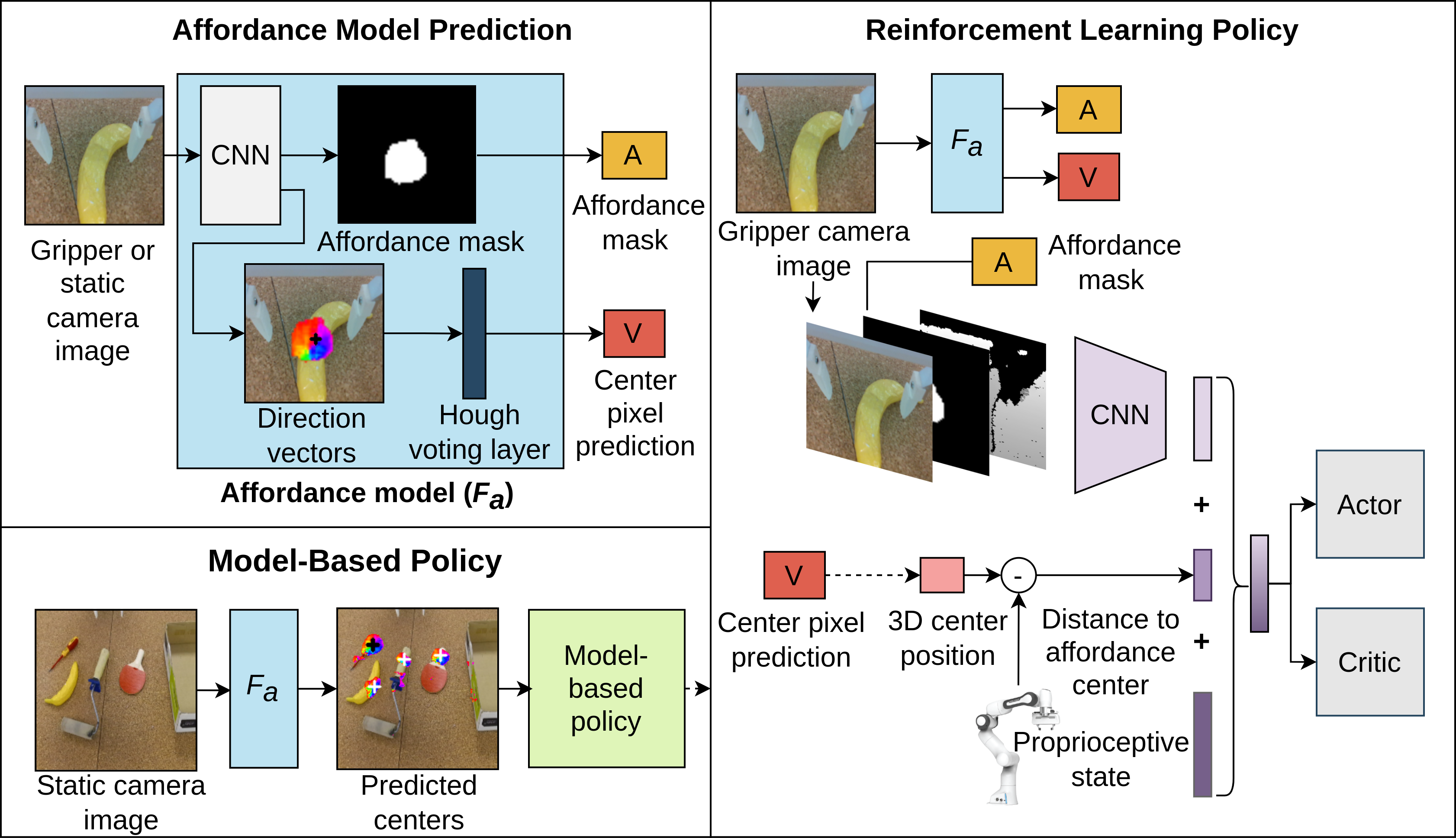}
  	\caption{Overview of the full approach. The affordance model takes an image from either camera as input to predict object affordance masks and center pixel predictions (top left). The static camera affordances are used to select a position that the model-based policy will move towards (bottom left). We then switch to a RL policy which takes as input the the predictions of the gripper camera affordance, the robot's proprioception, the distance to the predicted center, and the current RGB-D image
  	(right).}
	\label{fig:full_approach}
\end{figure}

We define a two-stage affordance detection by training separate models for the two cameras as shown in Figure~\ref{fig:full_approach}. One model is trained with images from a static camera and predicts a spatial interaction hotspots map, indicating actionable regions. Similarly, we train an affordance model with images from a gripper camera, which gives a finer-grained spatial interaction map about where humans tend to interact with each object.

The affordance model should give insight into which parts of an object are relevant for its use. As this is dependent on the shape of the objects rather than the color, we would like the affordance model to be invariant to different colors. For this reason, the images are converted to grayscale before being fed to the networks. 
Both affordance models are trained with stochastic gradient descent with a learning rate of 1e-5 and a batch size of 256. The loss weights are set to $w_{ce}=1$, $w_{dice}=5$, $w_{dir}=2.5$.

\textbf{Affordance-guided Reinforcement Learning}

We train the policy using Soft Actor-Critic \cite{sac}. We concatenate the RGB-D images with the inferred affordance mask and pass it through a convolutional neural network (CNN) as depicted in Figure~\ref{fig:full_approach}. The CNN is composed by three convolutional layers with kernel size [8,4,3] respectively and one linear layer to obtain a feature representation of size 16. Then we concatenate the obtained representation to the robot state and the distance to the affordance center. Finally this is passed through four fully connected layers. The critic and actor are implemented following the same architecture without weight-sharing.
For the simulation experiments, we train a single policy for all the objects with an episode length of 100 steps during 400K episode steps. This amounts to  30hrs of learning experience. We train for 3 seeds initializations. In the reward function we set $\lambda_1=\lambda_2=\lambda_3=1$ and the rewards $R_{succ}=200$, $R_{out}=-1$. 

\section{Experimental Results}
In this section we seek to answer the following questions: how does our method compare to the baseline policies in terms of sample efficiency and task completion? And, is the proposed approach applicable to a real world tidy-up task?

\subsection{Experimental Setup}
We evaluate our method with  both diverse simulation manipulation tasks and real world robot tidy-up experiments.
\subsubsection{Simulation}
\begin{figure}
    \centering
    \begin{subfigure}[b]{0.45\linewidth}
         \centering
         \includegraphics[width=\linewidth]{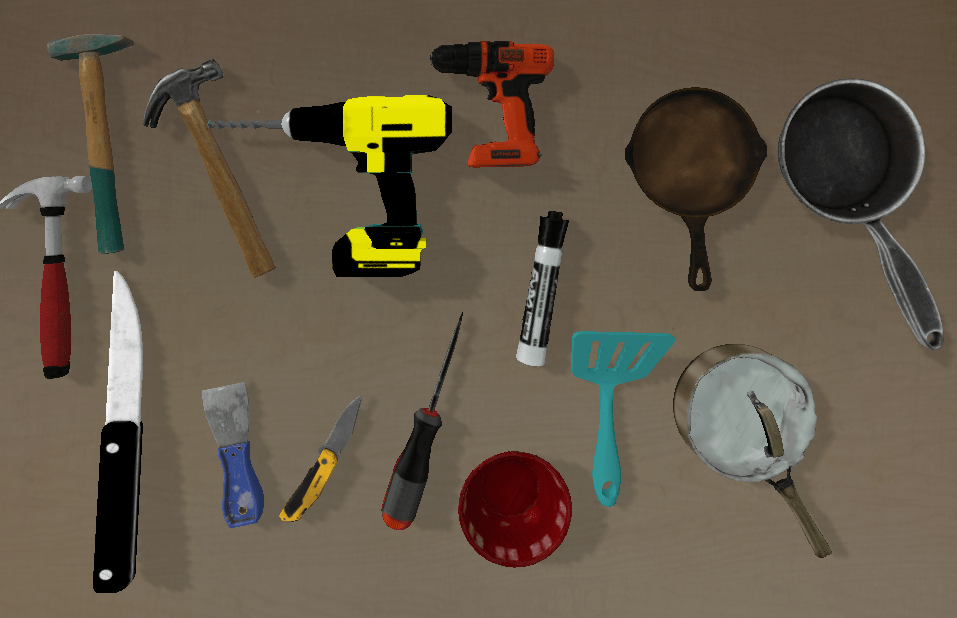}
         \caption{Seen objects during affordance model training.}
     \end{subfigure}
     \begin{subfigure}[b]{0.45\linewidth}
         \centering
         \includegraphics[width=\linewidth]{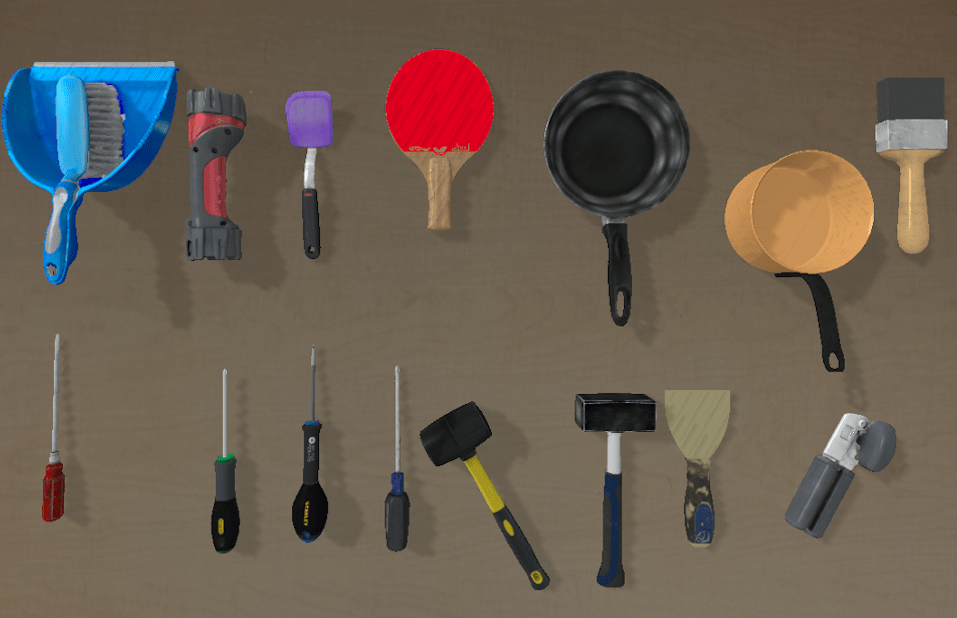}
         \caption{\textbf{Unseen} objects during affordance model training.}
     \end{subfigure}
    \caption{Objects used in simulation grasping experiments. The objects propose different challenges as some are small or must be grasped in a specific manner, e.g. grasping the frying pan requires to use the handle to be successfully lifted.}
    \label{fig:simulation_objects}
\end{figure}

We evaluate two tasks in simulation: a grasping task and a drawer opening task. The grasping task consists on lifting different objects in a PyBullet simulated environment. The policy is trained over 15 different objects with varying degrees of complexity, such as hammers, knifes and power drills, as shown in Figure~\ref{fig:simulation_objects}. After the policy executes a close-gripper action, the gripper attempts to lift the object and waits in the air for two seconds. If the object is still in the gripper at the end of this time, we define the grasp as being successful.

VAPO is not exclusive to a grasping task. To show this, we train a policy to open a drawer as shown in Figure \ref{fig:drawer}. Every episode consists of the drawer on a closed position and the robot in a neutral position. The episode is deemed successful if the robot opens the drawer at least 15cm.

To train the affordance models we teleoperate the robot using a virtual reality (VR) controller to collect unscripted play data. We gather  two hours of human interaction which amounts to $\sim$100K images for each environment to train the static camera and gripper camera affordance models.
\subsubsection{Real world}
For the real world experiment, we setup the environment using a 7-DOF Franka Emika Panda robot. 
The full setup can be seen in Figure~\ref{fig:real}. Similar as in simulation, we collect play data by teleoperating the robot using a VR controller as shown in Figure~\ref{fig:labeling}. We accumulate  1.5 hours of human interaction, which results in $\sim$70K images and use this to train both the gripper camera and static camera affordance models. 
The labels for both simulation and real world experiments are obtained as described in Section \ref{sec:affordance_play}. We only use the data to train the affordance models and do not need human annotation.

\subsection{Evaluation Protocol}
To test the sample efficiency of the affordance-guided RL policy, we compare against a  sparse-reward SAC agent, \emph{local-SAC}.  For this baseline, we remove $R_{aff}$ from the reward function and we modify the observation by removing the affordance mask and distance to the center. This policy still uses $\pi_{mod}$ to move through free space, but does not use the affordances for interaction. In essence, it is a sparse-reward SAC agent operating with the RGB-D images of the gripper camera in the vicinity of the objects. For all the experiments we show the success rate as the average success over a given number of attempts to complete a task.

\subsection{Simulation Experiments}

We start of by training policies to lift a diverse set of 15 objects on which the affordance model was also trained on. We observe that our approach outperforms the baseline and lifts significantly more objects as it has a strong prior on how objects should be interacted with. We observe that VAPO successfully can grasp objects at the anticipated afforded regions (handle of a pan, power-drill, knife), while the baseline fails to grasp objects of complex (frying pan) or ambiguous (bowl) geometries. This shows the effectiveness of the affordance-guided policy in learning stable functional grasps. Not only does our method learn better, but it is critically more sample-efficient. After $\sim$30 hours (400k timesteps) of robot interaction the baseline reaches a success rate of 0.6, while VAPO matches this performance at 100k steps. This indicates that our method learns up to $4 \times$ faster than the baseline. After training for 400k timesteps, VAPO remains stable at an overall success rate of 0.90.

Next we push our affordance model to generalize to unseen objects in two sets of experiments. In the first setting, we train and test the policies on 15 objects which were not seen by the affordance model during training.  We observe in Figure~\ref{fig:tabletop_episodes} that VAPO outperforms the baseline by a large margin in terms of both number of objects lifted and sample-efficiency. In the second setting, we evaluate the trained policies zero-shot on lifting 15 unseen objects. This
form of zero-shot evaluation is very challenging, as the objects are unseen for both the affordance model and the RL agent. We report a lifting success of 13/15 for VAPO and 8/15 for the baseline. This demonstrates the effectiveness of imbuing robots with an object-centric visual affordance. Aside from accelerating learning, the visual affordance model generalizes sufficiently to new object shapes and can anticipate their affordance regions, providing a useful object-centric prior. 

\begin{figure}[t]
    \centering
    \begin{tabular}{cc}
        \includegraphics[width=0.45\linewidth]{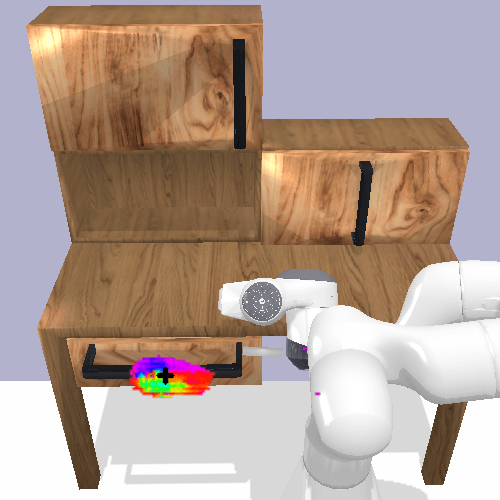} &
        \includegraphics[width=0.45\linewidth]{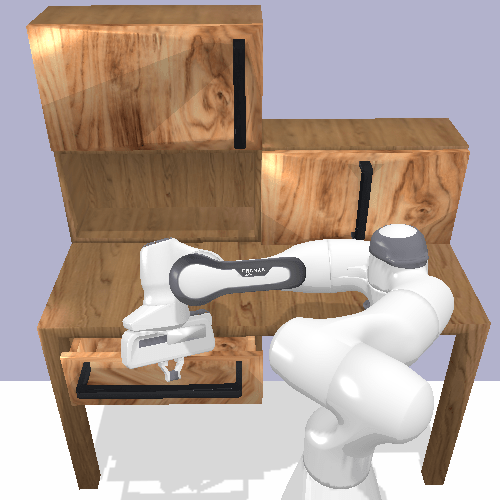} \\
    \end{tabular}
    \caption{Drawer opening task. On the left, the detected affordance and the corresponding center are shown. On the right we show a rollout of the RL agent opening the drawer.}
    \label{fig:drawer}
\end{figure}

\begin{figure}[b]
    \centering
     \begin{subfigure}[b]{0.46\linewidth}
         \centering
         \includegraphics[width=\linewidth]{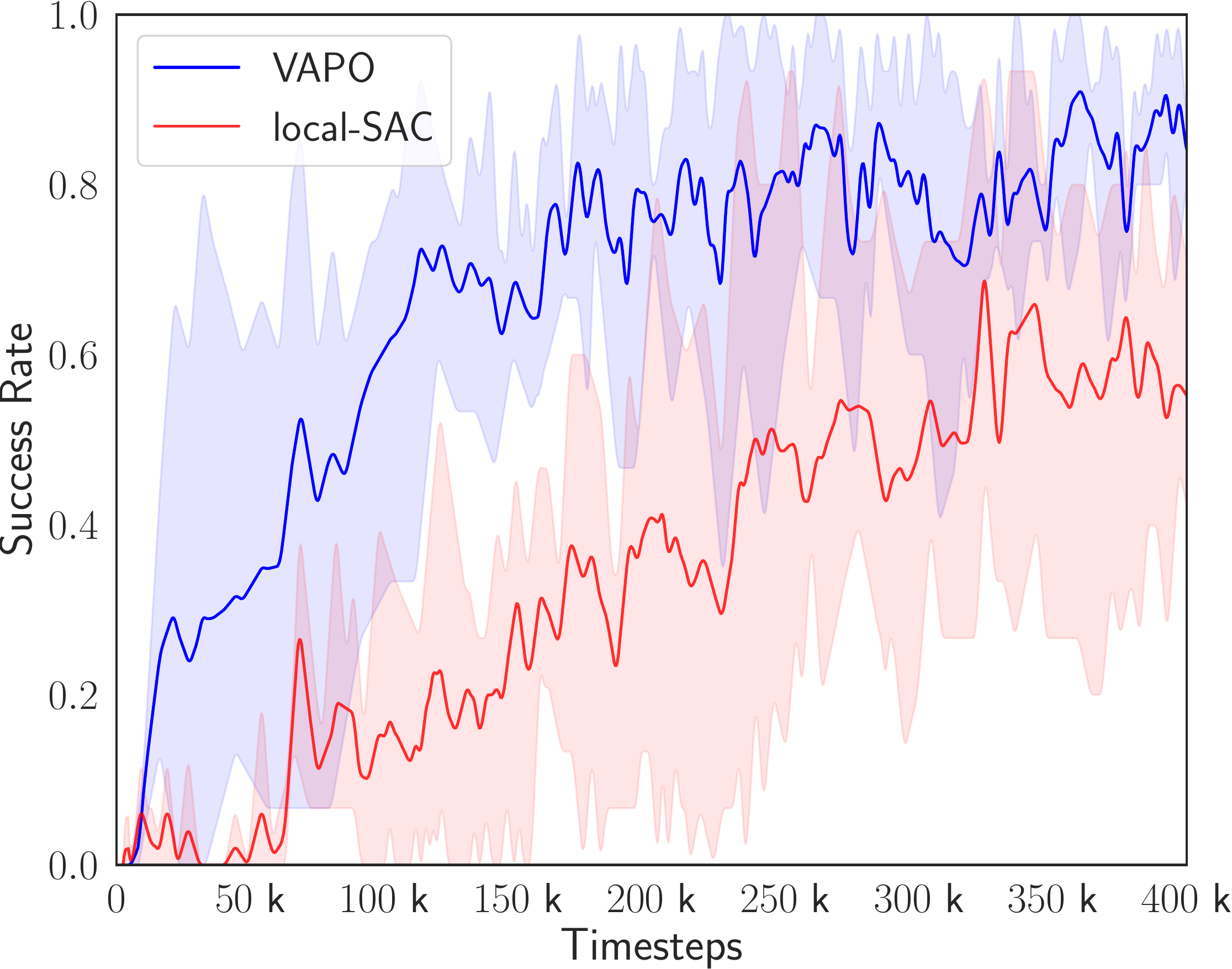}
         \caption{Policy trained on seen objects by the affordance model.}
         \label{fig:seen_results}
     \end{subfigure}
     \begin{subfigure}[b]{0.46\linewidth}
         \centering
         \includegraphics[width=\linewidth]{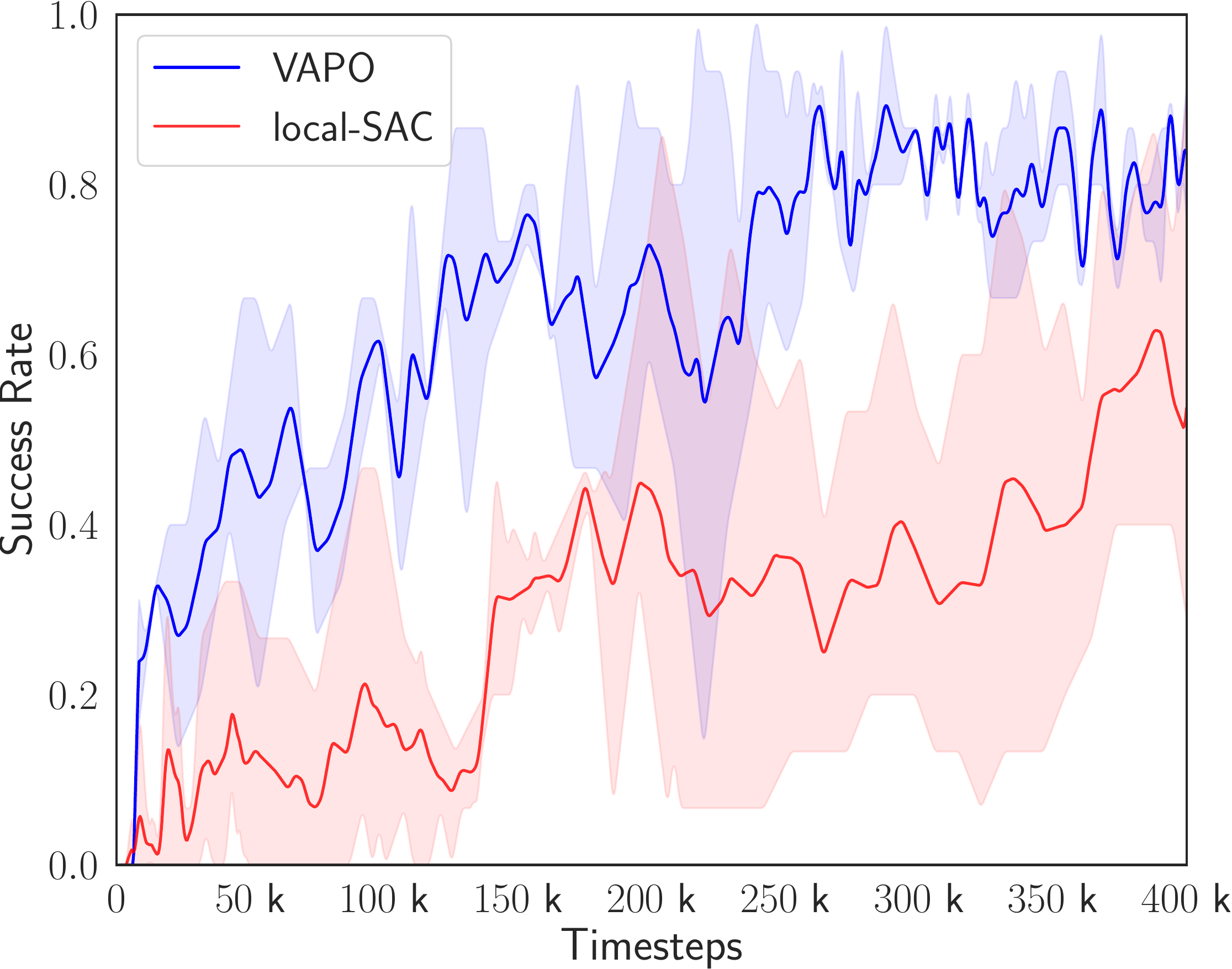}
         \caption{Policy trained on \textbf{unseen} objects by the affordance model.}
         \label{fig:unseen_results}
     \end{subfigure}
    \caption{VAPO  vs.~local-SAC for the pick up tasks. In both experiments, our method learns $4 \times$ faster as compared to the baseline and successfully lifts most of the objects. \label{fig:tabletop_episodes}}
 	\label{fig:tabletop_results}
\end{figure}

To analyze if our approach is applicable for more tasks, we conduct experiments on a drawer opening task (Figure \ref{fig:drawer}). We report a success rate over 100 episodes of 0.84 for VAPO and of 0.52 for the baseline. The results are consistent with the previous experiments showing that our method outperforms the baselines, while being more sample-efficient.

\subsection{Real World Experiments}
\begin{figure}[t]
    \centering
    \includegraphics[width=0.4\textwidth]{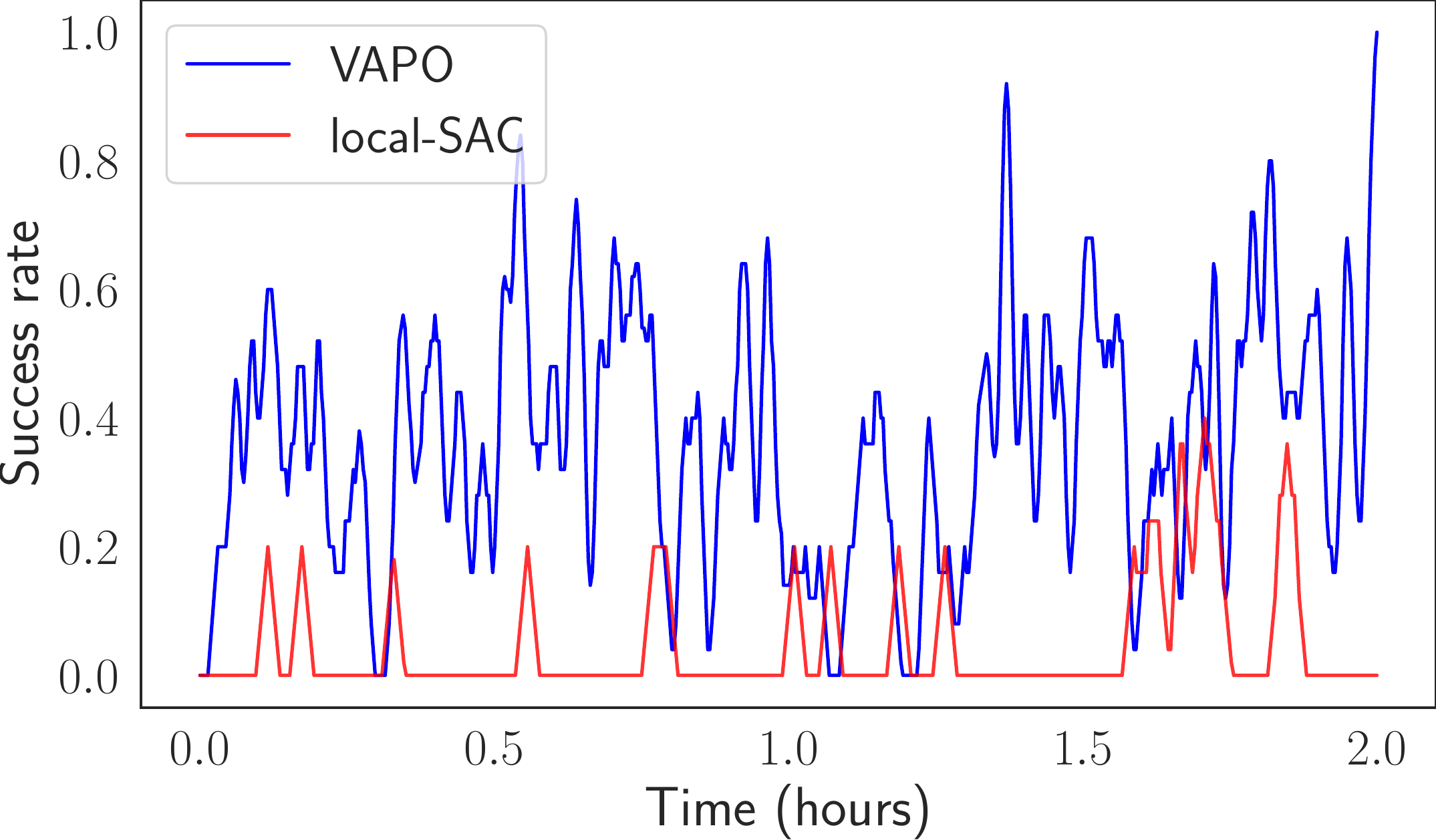}
    \caption{VAPO  vs.~local-SAC in real world tidy-up experiments. The success rate over the last ten episodes is shown. After two hours of real world robot interaction, the baseline rarely lifts any objects, while our approach 
    consistently ``functionally'' grasps all the objects.}
    \label{fig:real_world_curve}
\end{figure}

We finally evaluate our approach on a real world tidy-up experiment. We show the learning curves for this experiment in Figure~\ref{fig:real_world_curve}. We use a 7-DOF Franka Emika Panda robot and run our policy at 20 Hz. We train all methods to pickup four objects: a plastic banana, a screwdriver, a table tennis racket and a paint roller. After two hours of training VAPO is able to consistently ``functionally'' grasps all the objects, e.g., grasping the objects by the handles, while the SAC baseline rarely achieves to lift any object, despite the agent starting at the same robot pose as our method. This is due to the low number of samples that sparse-reward SAC is trained on, since most success stories of RL in the real world require several orders of magnitude more data~\cite{kalashnikov2018qt}. Overall, our results demonstrate the effectiveness of our approach to learn sample-efficient policies by leveraging self-supervised visual affordances.

\section{Conclusion}
\label{sec:conclusion}
In this paper, we introduced the novel approach VAPO (Visual Affordance-guided Policy Optimization) as a method for sample-efficient policy learning of manipulation tasks that is guided by a self-supervised visual affordance model.
The key advantage of our formulation is the extraction of visual affordances from unlabeled human teleoperated play data to learn a strong prior about  \emph{where}  actionable regions in an environment are. We distill this knowledge into an interplay between model-based and model-free policies that allows for a sample-efficient division of the robot control learning, without assuming a predefined set of manipulation primitives, 3D object shapes or a tracking system. 
Our results show that aside from accelerating learning, a critical advantage of imbuing robots with an object-centric visual affordance prior is the ability of policies to generalize to unseen, functionally similar, objects. To the best of our knowledge, this work is the first one to demonstrate the effectiveness of visual affordances to guide model-based policies and closed-loop RL policies to learn robot manipulation tasks in the real world. 







\bibliographystyle{IEEEtran}
\bibliography{root}

\end{document}